\documentclass[letterpaper, 10 pt, conference]{ieeeconf}

\IEEEoverridecommandlockouts
\usepackage{amsmath,amssymb,amsfonts}
\usepackage{algorithmic}
\usepackage{algorithm}
\usepackage{graphicx}
\usepackage{subcaption}
\usepackage{textcomp}
\usepackage{xcolor}
\usepackage{booktabs}
\usepackage{multirow}
\usepackage{hyperref}
\usepackage{subcaption}
\usepackage{tikz}
\usetikzlibrary{arrows.meta, positioning, fit, shapes.geometric, backgrounds, calc}

\newcommand{\xt}{\mathbf{x}_t}
\newcommand{\ut}{\mathbf{u}_t}
\newcommand{\gt}{\mathbf{g}_t}
\newcommand{\zt}{\mathbf{z}_t}
\newcommand{\ot}{\mathbf{o}_t}

\begin{document}

\bstctlcite{IEEEexample:BSTcontrol}

\title{\LARGE \bf
AeroBridge-TTA: Test-Time Adaptive Language-Conditioned Control for UAVs
}

\author{Lingxue Lyu*\\
University of Pennsylvania\\
Philadelphia, Pennsylvania, United States\\
{\tt\small lingxuelyu@alumni.upenn.edu}%
}

\maketitle
\thispagestyle{empty}
\pagestyle{empty}

\begin{abstract}
Language-guided unmanned aerial vehicles (UAVs) often fail not from bad reasoning or perception, but from \textit{execution mismatch}: the gap between a planned trajectory and the controller's ability to track it when the real dynamics differ from training (mass changes, drag shifts, actuator delay, wind). We propose \textbf{AeroBridge-TTA}, a language-conditioned control pipeline that targets this gap with test-time adaptation. It has three parts: a language encoder that maps the command into a subgoal, an adaptive policy conditioned on the subgoal and a learned latent, and a test-time adaptation (TTA) module that updates the latent online from observed transitions. On five language-conditioned UAV tasks under 13 mismatch conditions with the same domain randomization, AeroBridge-TTA ties a strong PPO-MLP baseline in-distribution and wins all 5 out-of-distribution (OOD) conditions, $+22.0$ pts on average ($62.7\%$ vs.\ $40.7\%$); the $+8.5$ pt overall gain comes entirely from the OOD regime. A same-weights ablation that only changes the step size $\alpha$ shows the latent update itself is responsible for a $4.6\times$ OOD lift. Code, checkpoints, and rollout videos: \url{https://github.com/lyulingxue/AeroBridge-TTA}.
\end{abstract}

\vspace{0.25em}
\noindent\textbf{\textit{Keywords---}}
Unmanned aerial vehicles, language-conditioned control, test-time adaptation, domain randomization, reinforcement learning, sim-to-real.
\vspace{0.25em}

\begin{figure}[t]
\centering
\includegraphics[width=\columnwidth]{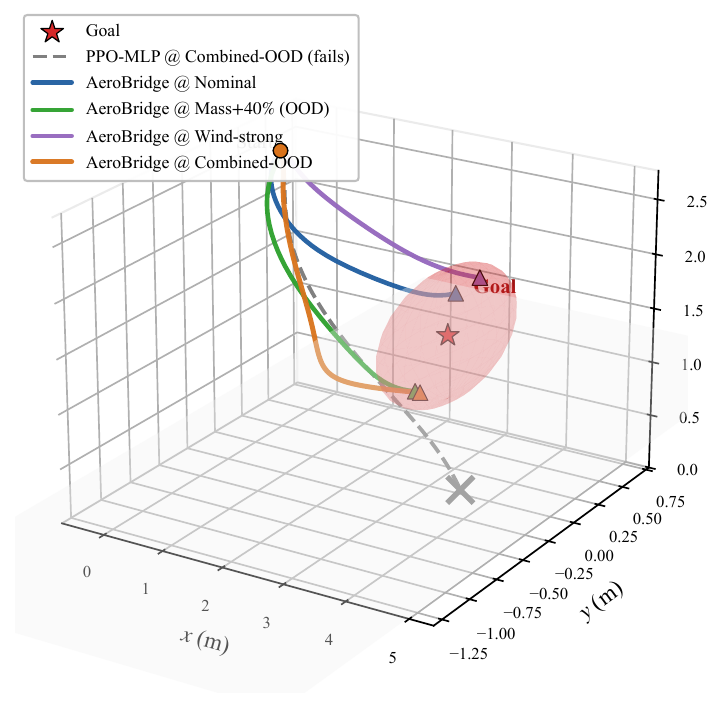}
\caption{\textbf{AeroBridge-TTA closes the execution-mismatch gap on unseen dynamics.} On the same language-conditioned navigation task, the PPO-MLP baseline fails under a composite out-of-distribution perturbation (gray, dashed, $\times$). The same AeroBridge-TTA checkpoint reaches the goal under nominal, mass$+40\%$ (OOD), strong wind, and combined-OOD, by adapting a latent from observed transitions online.}
\label{fig:teaser}
\end{figure}

\section{Introduction}
\label{sec:intro}

Large language models (LLMs) make it tempting to command robots in plain English instead of writing waypoints by hand~\cite{brohan2023rt2, driess2023palme, ahn2022saycan}. For unmanned aerial vehicles (UAVs) the appeal is even stronger: an operator can just say ``fly low to the target while avoiding the obstacle''~\cite{liu2023aerialvln, chen2024mapgpt}.

The reasoning and perception layers of these systems have improved a lot~\cite{vemprala2024chatgptrobotics, song2023llmplanner}, but one bottleneck is often ignored: \textbf{execution mismatch}. Even when the language module gives the right high-level plan, the low-level controller can fail when the real dynamics differ from training. Payload changes, damaged propellers, actuator delay, and wind all push the realised trajectory away from the intended one. The drone does not misunderstand the command; it just cannot fly the plan.

Online adaptation has a long history in other robot domains, and the lesson is consistent. In humanoid robotics, recovering balance from a sudden push needs millisecond-scale strategy switching. One line of work uses hand contact to stabilise a falling humanoid via real-time optimal control~\cite{wang2017realtime}, and later realises the same strategy in closed loop on a physical robot~\cite{wang2018realization}. More general planners pick the right contact mode from the available environment affordances by searching a contact transition tree~\cite{wang2018unified}, instead of committing to one fixed strategy. Legged locomotion uses a rapid motor adaptation module to handle new body parameters and terrain at deployment~\cite{kumar2021rma}. In agile flight, learned residual models adapt to wind in-flight~\cite{oconnell2022neural} or compensate ground effect during landing~\cite{shi2019neural}. Even a simple platform like a self-balancing bike needs the controller to react to the \emph{real} dynamics, not the assumed ones~\cite{sharma2016bicycle}; classical adaptive control formalises this idea with Lyapunov update laws~\cite{slotine1991applied}. The same message shows up in autonomous driving: small mismatches between a calibrated and the actual longitudinal model compound quickly and need online correction~\cite{wang2024calibration}. The common conclusion is that robust execution needs a closed-loop adaptation step \emph{inside} the controller, not just a better open-loop plan.

We propose \textbf{AeroBridge-TTA} for the language-conditioned UAV setting. Our contributions are:

\begin{enumerate}
    \item \textbf{Execution-mismatch hypothesis.} We name execution mismatch as a primary failure mode of language-conditioned UAV control, separate from perception or reasoning errors, and we show that domain randomisation alone does not close it (Sec.~\ref{sec:problem}).
    \item \textbf{Structured language-to-control pipeline.} A three-stage architecture, language encoder $\rightarrow$ subgoal $\rightarrow$ adaptive policy, that separates semantic understanding from motor execution. Each module can be improved on its own; the language frontend in particular is drop-in and can be replaced by a stronger LLM/VLA without retraining the controller (Sec.~\ref{sec:method}).
    \item \textbf{Test-time adaptation for UAV control.} A small TTA module that updates a latent $\zt$ from each observed transition, so the policy compensates dynamics mismatch \emph{without retraining}. The TTA head is around 6.5K parameters and runs as one feed-forward pass per step, suitable for 50\,Hz real-time control (Sec.~\ref{sec:tta}).
\end{enumerate}

We test AeroBridge-TTA on five language-conditioned tasks and 13 mismatch conditions, and compare it to a classical PID controller and to a strong PPO-MLP baseline trained with the same domain randomisation.

\section{Related Work}
\label{sec:related}

\textbf{Language-conditioned robot control.} SayCan~\cite{ahn2022saycan} grounds instructions in affordances; RT-2~\cite{brohan2023rt2} and PaLM-E~\cite{driess2023palme} train vision-language-action models end-to-end; LM-Nav~\cite{shah2023lmnav} reuses pre-trained models for outdoor navigation. Aerial work covers vision-and-language navigation~\cite{liu2023aerialvln} and LLM-based task planning~\cite{vemprala2024chatgptrobotics, chen2024mapgpt}. Most of these works assume the low-level controller is good enough, which is exactly what breaks under dynamics mismatch.

\textbf{UAV control and learning.} Classical work uses geometric controllers~\cite{lee2010geometric} and trajectory optimisation~\cite{mellinger2011minimum}. RL has been applied to quadrotors~\cite{hwangbo2017control} and to champion-level drone racing~\cite{kaufmann2023champion, loquercio2021learning}. Neural-Fly~\cite{oconnell2022neural} and Neural Lander~\cite{shi2019neural} learn residual dynamics, but they are not language-conditioned.

\textbf{Sim-to-real and adaptation.} Domain randomisation~\cite{tobin2017domain, peng2018simtoreal} trains over a dynamics distribution. RMA~\cite{kumar2021rma} learns an adaptation module for legged robots. Test-time training~\cite{sun2020tta} and self-supervised policy adaptation~\cite{hansen2021tta} update the model at deployment. Online calibration~\cite{wang2024calibration} corrects the dynamics for autonomous cars at test time. We bring this adaptation-at-execution idea to language-conditioned UAV control.

\textbf{Robust execution in robotics.} The need for closed-loop adaptation is well known in humanoids, which must recover from disturbances in real time~\cite{wang2017realtime, wang2018realization, wang2018unified}, and in balance problems~\cite{sharma2016bicycle}. Our take is the same: robustness under mismatch needs an adaptation step \emph{inside} the controller, not only a smarter plan.

\section{Problem Formulation}
\label{sec:problem}

\textbf{System.} We use a quadrotor with state $\xt = [\mathbf{p}, \mathbf{v}, \boldsymbol{\phi}, \boldsymbol{\omega}] \in \mathbb{R}^{12}$ (position, velocity, Euler angles, body rates) and input $\ut \in [-1,1]^4$ in collective-thrust-body-rates (CTBR) form. The dynamics are
\begin{align}
    m\dot{\mathbf{v}} &= R(\boldsymbol{\phi}) [0, 0, F_{\text{total}}]^{\top} - m\mathbf{g} + \mathbf{f}_{\text{drag}} + \mathbf{f}_{\text{wind}} \label{eq:trans} \\
    I\dot{\boldsymbol{\omega}} &= \boldsymbol{\tau} - \boldsymbol{\omega} \times (I\boldsymbol{\omega}). \label{eq:rot}
\end{align}

\textbf{Task.} Given a language command $q$, the controller must output a sequence $\{\ut\}_{t=0}^{T}$ that meets a per-task success check $\mathcal{S}_q(\xt)$ under reward $r_q(\xt, \ut)$.

\textbf{Execution mismatch.} We define it as the dynamics residual between training and deployment,
\begin{equation}
    \Delta f_t = f_{\text{test}}(\xt, \ut) - f_{\text{train}}(\xt, \ut),
    \label{eq:mismatch}
\end{equation}
with sources $\Delta m$, $\Delta c_d$, delay, wind. A policy that scores 100\% under $f_{\text{train}}$ can still fail under $f_{\text{test}}$. That is an execution problem, not a reasoning one.

\section{Method}
\label{sec:method}

AeroBridge-TTA consists of three components (Fig.~\ref{fig:architecture}):

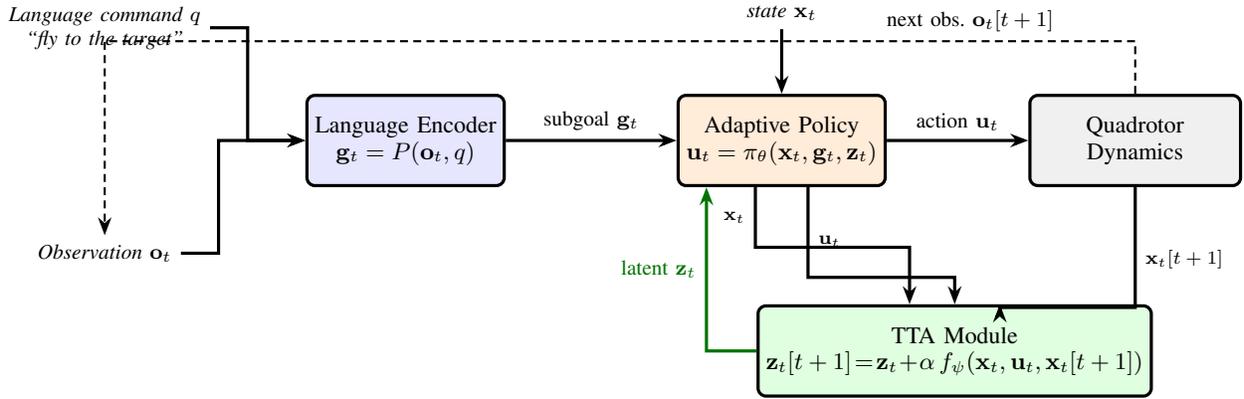
\begin{figure*}[t]
    \centering
    \begin{tikzpicture}[
        font=\small,
        >={Stealth[length=2.4mm,width=1.8mm]},
        module/.style={rectangle, rounded corners=3pt, draw, very thick,
                       minimum height=12mm, minimum width=26mm, align=center,
                       inner sep=3pt, font=\small},
        lang/.style={module, fill=blue!10},
        policy/.style={module, fill=orange!15},
        tta/.style={module, fill=green!12},
        env/.style={module, fill=gray!12, minimum width=28mm},
        io/.style={align=center, font=\footnotesize\itshape},
        signal/.style={very thick, ->},
        zarrow/.style={very thick, ->, color=green!45!black},
        feedback/.style={thick, densely dashed, ->},
    ]
        \node[io] (cmd) at (0, 1.5) {Language command $q$\\``fly to the target''};
        \node[io] (obs) at (0,-1.5) {Observation $\ot$};

        \node[lang] (lang) at (4.0, 0) {Language Encoder\\$\gt = P(\ot, q)$};

        \node[policy] (pol) at (9.0, 0) {Adaptive Policy\\$\ut = \pi_\theta(\xt,\gt,\zt)$};

        \node[env] (env) at (13.7, 0) {Quadrotor\\Dynamics};

        \node[io] (stlab) at (9.0, 1.7) {state $\xt$};
        \draw[signal] (stlab.south) -- (pol.north);

        \node[tta] (tta) at (11.3, -2.8) {TTA Module\\$\zt[t+1]\!=\!\zt\!+\!\alpha\, f_\psi(\xt,\ut,\xt[t+1])$};

        \draw[signal] (cmd.east)  -- ++(0.5,0) |- (lang.west);
        \draw[signal] (obs.east)  -- ++(0.5,0) |- (lang.west);

        \draw[signal] (lang.east) --
              node[above, font=\footnotesize]{subgoal $\gt$} (pol.west);

        \draw[signal] (pol.east) --
              node[above, font=\footnotesize]{action $\ut$} (env.west);

        \draw[feedback] (env.north) -- ++(0, 0.7) -| (obs.north)
              node[pos=0.08, above, font=\footnotesize]{next obs.\ $\ot[t+1]$};

        \draw[signal] ($(pol.south)+(-0.35,0)$) -- ++(0,-0.8)
              node[midway, left, font=\scriptsize]{$\xt$}
              -| ($(tta.north)+(-0.6,0)$);
        \draw[signal] ($(pol.south)+(0.35,0)$) -- ++(0,-1.2)
              node[pos=0.6, right, font=\scriptsize]{$\ut$}
              -| (tta.north);
        \draw[signal] (env.south) -- ++(0, -1.6)
              node[pos=0.6, right, font=\scriptsize]{$\xt[t+1]$}
              -| ($(tta.north)+(0.6,0)$);

        \draw[zarrow] (tta.west) -|
              node[pos=0.75, left, font=\footnotesize]{latent $\zt$}
              ($(pol.south)+(-1.0,0)$);
    \end{tikzpicture}
    \caption{AeroBridge-TTA architecture. The language encoder $P$ maps the command $q$ and observation $\ot$ to a subgoal $\gt$; the adaptive policy $\pi_\theta$ generates actions $\ut$ conditioned on the state $\xt$, subgoal $\gt$, and adaptation latent $\zt$. The TTA module $f_\psi$ updates $\zt$ online from the observed transition $(\xt,\ut,\xt[t+1])$. Dashed arrows are environment feedback; the green arrow highlights the test-time adaptation loop.}
    \label{fig:architecture}
\end{figure*}

\textbf{Language encoder $\gt = P(\ot, q)$.} It maps the command $q$ and observation $\ot$ to a position-delta subgoal $\gt\in\mathbb{R}^3$:
\begin{equation}
    \gt = \text{MLP}([\text{Embed}(q); \ot]),
    \label{eq:subgoal}
\end{equation}
where $\text{Embed}(q)\in\mathbb{R}^{32}$ is a learned task embedding and $\ot\in\mathbb{R}^{32}$ is the full observation (state, relative goal, task one-hot, obstacle info). The subgoal separates \emph{what to do} from \emph{how to do it}.

\begin{figure}[t]
    \centering
    \includegraphics[width=\columnwidth]{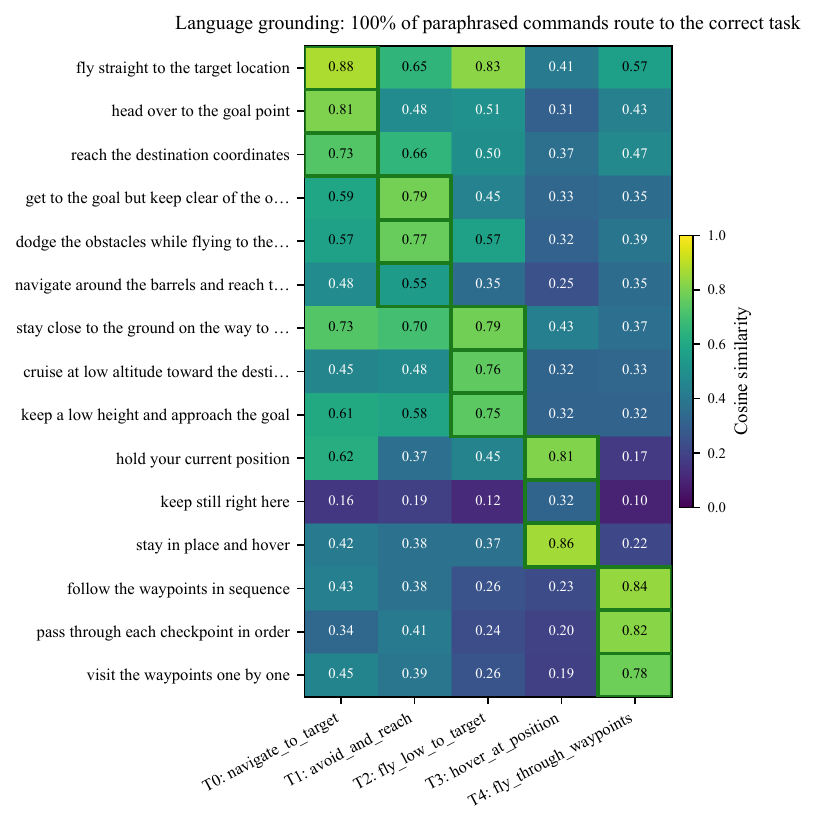}
    \caption{\textbf{Language grounding POC.} Cosine similarity between $15$ free-form commands (rows) and $5$ canonical templates (columns), with MiniLM-L6-v2 and per-task max-pooling. Argmax per row is boxed; all $15$ route correctly.}
    \label{fig:language_grounding}
\end{figure}

\textbf{Language grounding frontend (POC).} During training $q$ is a one-hot, but a real operator types free-form English. We bridge the two with a frozen sentence encoder (MiniLM-L6-v2, $384$-d~\cite{reimers2019sbert}). For each task $k$ we precompute embeddings $\phi(\mathbf{s})$ for a small bundle of canonical paraphrases $\mathcal{S}_k$, and route an incoming command $q_{\text{nl}}$ by
\begin{equation}
    \hat{q} = \arg\max_{k}\; \max_{\mathbf{s} \in \mathcal{S}_k}\; \cos\!\bigl(\phi(q_{\text{nl}}),\, \phi(\mathbf{s})\bigr).
    \label{eq:grounding}
\end{equation}
The predicted $\hat{q}$ feeds Eq.~\ref{eq:subgoal}. The frontend is drop-in: a stronger LLM or VLA (GPT-4, PaLM-E, RT-2) can replace $\phi$ without touching the controller. On $15$ held-out paraphrases that share no lexical overlap with the templates, the grounder reaches $100\%$ task-routing accuracy (Fig.~\ref{fig:language_grounding}).

\textbf{Adaptive policy $\ut = \pi_\theta(\xt, \gt, \zt)$.} The policy takes state, subgoal, and adaptation latent:
\begin{equation}
    \ut \sim \mathcal{N}\!\left(\tanh\!\left(\text{MLP}_\theta([\xt; \gt; \zt])\right),\, \text{diag}(\boldsymbol{\sigma}_\theta^2)\right),
    \label{eq:policy}
\end{equation}
with three 256-unit $\tanh$ layers, orthogonal init, learnable log-std at $e^{-1.5}$. Conditioning on $\zt$ lets the policy react to the inferred mismatch without an explicit system identification step.

\textbf{Test-time adaptation module.}
\label{sec:tta}
A latent $\zt \in \mathbb{R}^{32}$ is updated each step from the observed transition:
\begin{equation}
    \zt[t+1] = \zt + \alpha \cdot f_\psi(\xt, \ut, \xt[t+1]),
    \label{eq:tta}
\end{equation}
with $f_\psi$ a small $\tanh$-output MLP ($\mathbb{R}^{28}\!\to\!\mathbb{R}^{32}$, about 6.5K params) and $\alpha=0.1$. The residual carries the mismatch signal: $f_\psi$ is trained end-to-end with PPO~\cite{schulman2017ppo} under DR (mass $\in[0.9,1.1]$, drag $\in[0.8,1.2]$) so that $\zt$ becomes useful to the policy. A separate value head $V_\phi(\ot)$ is trained jointly. At deployment the update runs online with no gradient steps.

\section{Experiments}
\label{sec:experiments}

We ask three questions: (1) can AeroBridge-TTA do all five language-conditioned tasks? (2) does TTA help under mismatch? (3) is the gain from the adaptation mechanism, not just from extra parameters?

\begin{figure}[t]
\centering
\begin{subfigure}[b]{\columnwidth}
    \centering
    \includegraphics[width=\textwidth]{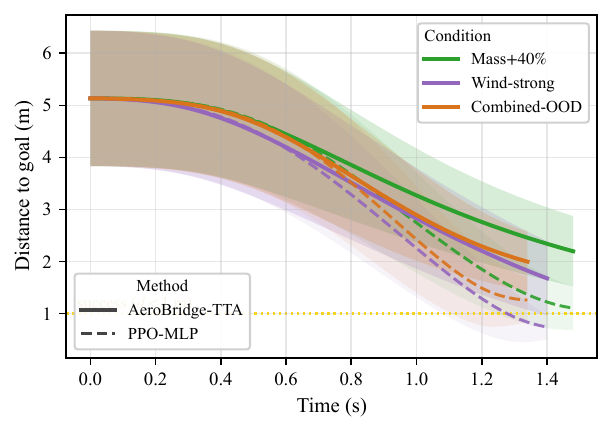}
    \caption{Distance to goal across 3 OOD conditions (Mass$+40\%$, Wind-strong, Combined-OOD), mean $\pm 1\sigma$ over 10 seeds. AeroBridge-TTA (solid) drives the distance below the $1$\,m success threshold; the PPO-MLP baseline (dashed) saturates at a large standoff on the harder perturbations.}
    \label{fig:sim_traj_dist}
\end{subfigure}

\vspace{0.4em}

\begin{subfigure}[b]{\columnwidth}
    \centering
    \includegraphics[width=\textwidth]{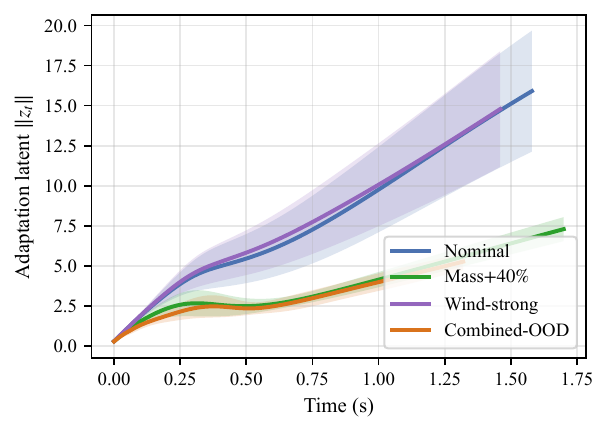}
    \caption{Adaptation latent $\Vert z_t\Vert$ for AeroBridge-TTA across nominal and 3 OOD conditions, mean $\pm 1\sigma$ over 10 seeds. Under nominal dynamics the latent stays close to zero; under OOD perturbations the TTA module grows the latent, with magnitude scaling with perturbation severity.}
    \label{fig:sim_traj_z}
\end{subfigure}
\caption{Time-series diagnostics from the trained AeroBridge-TTA checkpoint and the PPO-MLP+DR baseline. The 3D trajectory view is in Fig.~\ref{fig:teaser}.}
\label{fig:sim_traj}
\end{figure}

\textbf{Environment.} A 6-DOF quadrotor sim implementing Eqs.~\ref{eq:trans}--\ref{eq:rot} with aerodynamic drag and configurable wind, running at 50\,Hz. State $\in\mathbb{R}^{12}$, action $\in[-1,1]^4$ (CTBR), observation $\in\mathbb{R}^{32}$. The five tasks are listed in Table~\ref{tab:tasks}. Teaser trajectories are in Fig.~\ref{fig:teaser}; distance and latent time-series are in Fig.~\ref{fig:sim_traj}.

\begin{table}[t]
\centering
\caption{Language-conditioned task definitions.}
\label{tab:tasks}
\resizebox{\columnwidth}{!}{%
\begin{tabular}{@{}clc@{}}
\toprule
ID & Language Command & Success Criterion \\
\midrule
0 & ``Go to the target position''     & $d\!<\!1.0$\,m, $\|\mathbf{v}\|\!<\!2.0$\,m/s \\
1 & ``Avoid obstacle, reach target''  & $d\!<\!1.0$\,m, $\|\mathbf{v}\|\!<\!2.0$, no collision \\
2 & ``Fly low and reach target''      & $d\!<\!1.0$\,m, $\|\mathbf{v}\|\!<\!2.0$, $p_z\!<\!1.0$\,m \\
3 & ``Hover at the designated pose''  & $d\!<\!0.5$\,m, $\|\mathbf{v}\|\!<\!0.3$\,m/s \\
4 & ``Fly through waypoints in order''& all 3 WPs: $d\!<\!1.0$\,m, $\|\mathbf{v}\|\!<\!2.0$ \\
\bottomrule
\end{tabular}%
}
\end{table}

\textbf{Baselines.} PID (cascaded position/attitude~\cite{lee2010geometric}), PPO-MLP (3-layer MLP with full obs including task one-hot)~\cite{schulman2017ppo}, Ours without TTA ($\zt\equiv\mathbf{0}$), and full Ours.

\textbf{Training.} PPO ($\gamma{=}0.99$, $\lambda_{\text{GAE}}{=}0.95$, clip $0.2$, lr $3\!\times\!10^{-4}$ linear, entropy $10^{-3}$), 32 parallel envs, 4096 steps per rollout. A distance curriculum scales goals from 2\,m to 7\,m (+0.15\,m per step above $25\%$ SR, 10-iter EMA). Reward follows the profile $v^\star{=}\min(3.5, 1.5\sqrt{d})$ plus a $+500$ terminal bonus. Training converges in about 50 iters (about 6.5M steps) on an RTX 5080.

\subsection{Experiment 1: Nominal Task Performance}

First we ask whether the AeroBridge architecture can do all five tasks under nominal dynamics. Table~\ref{tab:task_results} shows three settings: (1) per-task PPO-MLP, (2) per-task AeroBridge without TTA, and (3) a \textit{single} multi-task AeroBridge model trained on all five tasks at once with language conditioning.

\begin{table}[t]
\centering
\caption{Success rate (\%) across language-conditioned tasks under nominal dynamics (50 evaluation episodes). ``Multi-task'' uses a \textit{single} language-conditioned model for all tasks.}
\label{tab:task_results}
\resizebox{\columnwidth}{!}{%
\begin{tabular}{@{}lcccccc@{}}
\toprule
Method & Nav. & Avoid & Low & Hover & WP & Avg. \\
\midrule
PPO-MLP (per-task)         & 100  & --   & --  & --  & --  & --   \\
Ours w/o TTA (per-task)    & 100  & 98   & 100 & 100 & 100 & 99.6 \\
\textbf{Ours (multi-task)} & \textbf{100}  & \textbf{88}   & \textbf{100} & \textbf{100} & \textbf{100} & \textbf{97.6} \\
\bottomrule
\end{tabular}%
}
\vspace{-1em}
\end{table}

A \emph{single} AeroBridge model, conditioned only on the language embedding, reaches $97.6\%$ average SR over all five tasks ($100/88/100/100/100$). PPO-MLP needs five separate per-task models to do the same. The multi-task run uses 40 parallel envs (8 per task) and a per-task curriculum (hover starts at $40\%$ of the nav distance because its success criterion is tighter), and converges in about 35 iters (about 5.7M steps). The drop on the avoidance task ($88\%$) comes from a small set of obstacle layouts in which the multi-task model still skims the corner; per-task training closes this gap. The point of this experiment is not to push every task to $100\%$, but to show that one language-conditioned controller can cover all five.

\subsection{Experiment 2: Dynamics Mismatch Robustness}

To test robustness, we train PPO-MLP and AeroBridge-TTA from scratch on the navigation task (Task~0) with domain randomization (DR): mass uniform in $[0.8, 1.3]\times$ and drag in $[0.5, 2.0]\times$ at every episode reset. Both run for about 13M steps. We then evaluate them on 13 mismatch conditions (30 episodes each). Two conditions are strictly out-of-distribution for the DR mass range (Mass$+40\%$ and Combined-OOD, marked $^\ast$). Wind and control delay are not randomized during training, so those axes are also untrained. We group all of these as the OOD set below.

\begin{table}[t]
\centering
\caption{Mismatch conditions ($^\ast$ = OOD relative to the DR training range).}
\label{tab:mismatch_conditions}
\small
\begin{tabular}{@{}lcccc@{}}
\toprule
Condition & Mass & Drag & Delay & Wind (N) \\
\midrule
Nominal             & 1.0$\times$ & 1.0$\times$ & 0 & -- \\
Mass$-20\%$         & 0.8$\times$ & 1.0$\times$ & 0 & -- \\
Mass$+20\%$         & 1.2$\times$ & 1.0$\times$ & 0 & -- \\
Mass$+30\%$         & 1.3$\times$ & 1.0$\times$ & 0 & -- \\
Mass$+40\%^\ast$    & 1.4$\times$ & 1.0$\times$ & 0 & -- \\
Drag$+100\%$        & 1.0$\times$ & 2.0$\times$ & 0 & -- \\
Delay (2)           & 1.0$\times$ & 1.0$\times$ & 2 & -- \\
Delay (5)           & 1.0$\times$ & 1.0$\times$ & 5 & -- \\
Wind med.           & 1.0$\times$ & 1.0$\times$ & 0 & [1.0, 0.5, 0] \\
Wind strong         & 1.0$\times$ & 1.0$\times$ & 0 & [2.0, 1.0, 0.3] \\
Combined mild       & 1.1$\times$ & 1.3$\times$ & 1 & [0.3, 0.1, 0] \\
Combined hard       & 1.2$\times$ & 1.5$\times$ & 3 & [1.0, 0.5, 0] \\
Combined-OOD$^\ast$ & 1.4$\times$ & 1.8$\times$ & 3 & [1.5, 0.8, 0.2] \\
\bottomrule
\end{tabular}
\vspace{-0.5em}
\end{table}

Table~\ref{tab:mismatch_results} compares the two methods. The picture is clearly asymmetric. In-distribution they are tied ($96.3\%$ vs.\ $96.2\%$, $\Delta$ID $=+0.0$). Out-of-distribution, AeroBridge-TTA wins every single condition, $+22.0$ pts on average ($62.7\%$ vs.\ $40.7\%$). Across all 13 conditions the gain is $+8.5$ pts ($83.3\%$ vs.\ $74.9\%$), and it comes entirely from the OOD side.

\begin{table}[t]
\centering
\caption{Success rate (\%) under dynamics mismatch (Task 0: Navigate, 60 episodes per condition). Both methods use identical domain randomization during training. Conditions are ordered in-distribution (ID) first, then out-of-distribution (OOD). Best per condition in \textbf{bold}; $^\ast$ marks strict OOD conditions beyond the DR mass range.}
\label{tab:mismatch_results}
\resizebox{\columnwidth}{!}{%
\begin{tabular}{@{}lccc@{}}
\toprule
Condition & PPO-MLP + DR & AeroBridge-TTA + DR & $\Delta$ \\
\midrule
\multicolumn{4}{@{}l}{\emph{In-distribution (DR covers the perturbation)}} \\
Nominal         & \textbf{100} & \textbf{100} & \phantom{0}$0$ \\
Mass$-20\%$     & \textbf{100} & \phantom{0}91.7 & \phantom{0}$-8.3$ \\
Mass$+20\%$     & \phantom{0}90 & \textbf{\phantom{0}96.7} & \phantom{0}$+6.7$ \\
Mass$+30\%$     & \phantom{0}80 & \textbf{\phantom{0}91.7} & $+11.7$ \\
Drag$+100\%$    & \textbf{100} & \phantom{0}98.3 & \phantom{0}$-1.7$ \\
Delay (2)       & \textbf{100} & \textbf{100} & \phantom{0}$0$ \\
Wind med.       & \textbf{100} & \phantom{0}96.7 & \phantom{0}$-3.3$ \\
Combined mild   & \textbf{100} & \phantom{0}95.0 & \phantom{0}$-5.0$ \\
\emph{ID avg}    & \emph{96.2}  & \textbf{\emph{96.3}}  & \emph{$+0.0$} \\
\midrule
\multicolumn{4}{@{}l}{\emph{Out-of-distribution (beyond DR or untrained axis)}} \\
Mass$+40\%^\ast$      & 23.3 & \textbf{\phantom{0}56.7} & $+33.3$ \\
Wind strong           & 95.0 & \textbf{\phantom{0}96.7} & \phantom{0}$+1.7$ \\
Combined hard         & 51.7 & \textbf{\phantom{0}81.7} & $+30.0$ \\
Delay (5)             & 28.3 & \textbf{\phantom{0}40.0} & $+11.7$ \\
Combined-OOD$^\ast$   & \phantom{0}5.0 & \textbf{\phantom{0}38.3} & $+33.3$ \\
\emph{OOD avg}        & \emph{40.7} & \textbf{\emph{62.7}} & \emph{$+22.0$} \\
\midrule
\textbf{Overall average} & 74.9 & \textbf{83.3} & \textbf{$+8.5$} \\
\bottomrule
\end{tabular}%
}
\vspace{-1em}
\end{table}

\begin{figure}[t]
\centering
\includegraphics[width=\columnwidth]{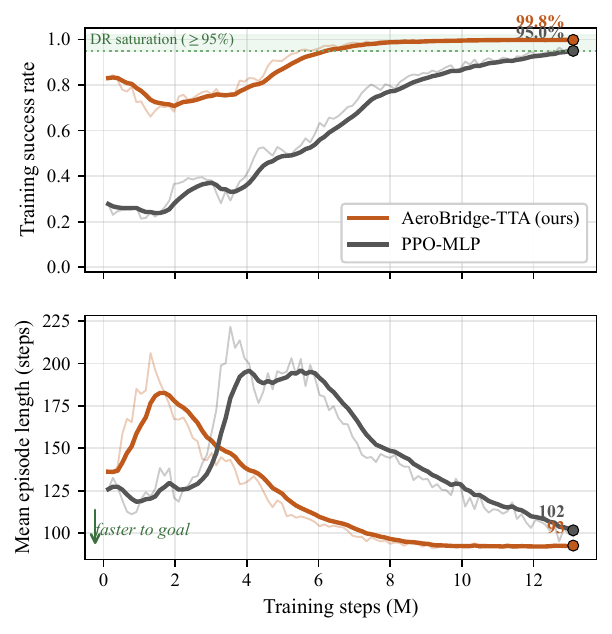}
\caption{Training behaviour under domain randomization for AeroBridge-TTA (ours) and PPO-MLP. \textbf{Top:} both methods reach $\geq 95\%$ training SR on the DR distribution (AB $100\%$, BL $\sim95\%$); the OOD gap in Table~\ref{tab:mismatch_results} therefore comes from adaptation at test time, not from training saturation. \textbf{Bottom:} mean episode length --- AeroBridge-TTA finishes trajectories in $\sim93$ steps vs.\ $\sim100$ for the baseline, i.e.\ reaches the goal $\sim7\%$ faster even on the same task. Raw traces in light shading; EMA-smoothed primary in bold.}
\label{fig:training_dr}
\end{figure}

\textbf{In-distribution: tied.} Over the 8 ID conditions the two methods are basically the same on average ($96.3\%$ vs.\ $96.2\%$). Already at the DR edge AeroBridge-TTA is ahead ($+6.7$ on Mass$+20\%$, $+11.7$ on Mass$+30\%$), which already hints at the OOD story. In training it also reaches the goal about $7\%$ faster (about 93 vs.\ 100 steps, Fig.~\ref{fig:training_dr}).

\textbf{OOD: AeroBridge-TTA wins every condition.} Outside DR coverage, our method wins all 5 OOD conditions: $+33.3$ on Mass$+40\%^\ast$ ($2.4\times$), $+1.7$ on Wind-strong, $+11.7$ on Delay$=5$, $+30.0$ on Combined-hard, $+33.3$ on Combined-OOD$^\ast$ ($7.7\times$). The biggest gains are on axes that DR did not see (Delay, Wind) or that stack multiplicatively (Combined). One latent summarising ``what is off'' is more efficient than expanding the DR grid. (We exclude Mass$+50\%$: both score $0\%$ because of the thrust ceiling.)

\subsection{Experiment 3: TTA Ablation}

Experiment 2 compared two different agents. Here we isolate the latent by running the \emph{same} trained AeroBridge-TTA model with three step sizes at test time: $\alpha = 0$ (no adaptation, $\zt \equiv \mathbf{0}$), $\alpha = 0.02$ (slow), and $\alpha = 0.1$ (full). Four mismatch conditions, 30 episodes each, covering both ID and OOD.

\begin{table}[t]
\centering
\caption{TTA ablation: success rate (\%) on the same trained AeroBridge-TTA checkpoint, varying only the adaptation step size $\alpha$ at test time (30 episodes per condition). $^\ast$ denotes strict OOD conditions.}
\label{tab:tta_ablation}
\resizebox{\columnwidth}{!}{%
\begin{tabular}{@{}lccc@{}}
\toprule
Condition & $\alpha{=}0$ (no TTA) & $\alpha{=}0.02$ (slow) & $\alpha{=}0.1$ (full) \\
\midrule
Mass$+30\%$          & 20 & 50 & \textbf{97} \\
Mass$+40\%^\ast$     & \phantom{0}0 & \phantom{0}3 & \textbf{57} \\
Combined-hard        & 40 & 60 & \textbf{93} \\
Combined-OOD$^\ast$  & \phantom{0}7 & \phantom{0}7 & \textbf{63} \\
\midrule
\textbf{Average} & 16.8 & 30.0 & \textbf{77.5} \\
\bottomrule
\end{tabular}%
}
\vspace{-0.5em}
\end{table}

\begin{figure}[t]
\centering
\includegraphics[width=\columnwidth]{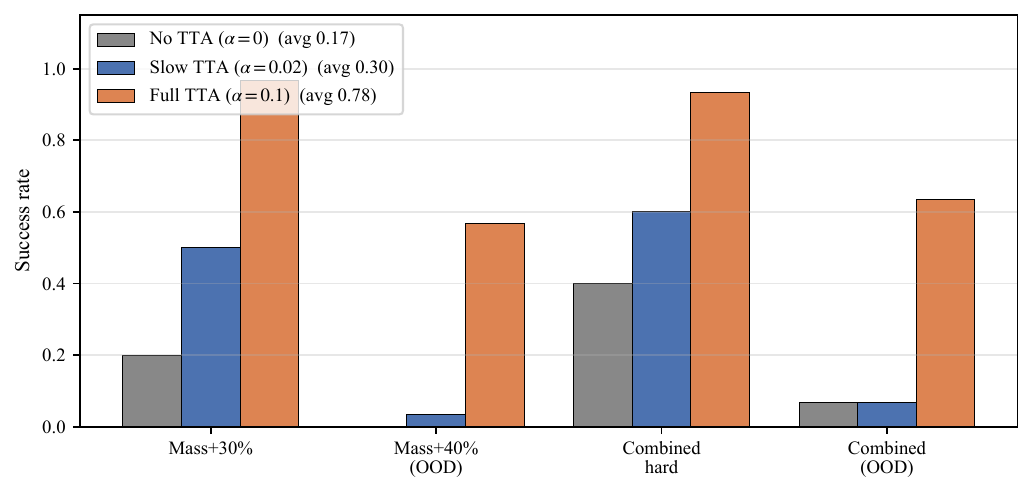}
\caption{TTA ablation on a single trained checkpoint. Full adaptation ($\alpha{=}0.1$) lifts average success from 16.8\% ($\alpha{=}0$) to 77.5\%, a $4.6\times$ improvement from the test-time latent update alone (weights identical across all three variants); slow adaptation ($\alpha{=}0.02$) is an intermediate compromise.}
\label{fig:tta_ablation}
\end{figure}

Averaged over the four conditions, $\alpha{=}0.1$ lifts SR from $16.8\%$ ($\alpha{=}0$) to $77.5\%$, a $4.6\times$ gain from only the test-time update, with the \emph{same weights}. The trend is monotone per condition ($20\!\to\!50\!\to\!97$ on Mass$+30\%$; $0\!\to\!3\!\to\!57$ on Mass$+40\%^\ast$; $40\!\to\!60\!\to\!93$ on Combined-hard; $7\!\to\!7\!\to\!63$ on Combined-OOD, a $9\times$ gain). Since only $\alpha$ changes, the OOD gain in Exp.~2 must come from the adaptation mechanism itself, not from extra parameters.

\section{Discussion}
\label{sec:discussion}

\textbf{DR and TTA are complementary.} Inside the DR range, both methods are saturated and basically tied. The gap opens on OOD, where AeroBridge-TTA wins every condition (up to $7.7\times$ on Combined-OOD). A latent that summarises the recent transition residuals is the right tool for perturbations DR cannot enumerate, especially when several axes (mass, drag, delay, wind) interact at once. Pushing the DR range wider would help in principle, but the effective grid grows multiplicatively with the number of axes, while one TTA latent costs us a single small MLP.

\textbf{Why the Combined-OOD gap is so large.} Stacking mass, drag, delay, and wind together creates a regime that the baseline never sees, even with wide DR. AeroBridge-TTA does not need to have seen this exact combination at training time: as long as the residual signature is in the span of what $f_\psi$ learned, the latent moves in a direction the policy already knows how to react to. This is why we observe a $7.7\times$ gain on Combined-OOD, while Wind-strong (a single, mild axis) gives a much smaller gain.

\textbf{Limits.} Adaptation cannot beat physical limits: at Mass$+50\%$ both methods fail because the rotors run out of thrust. A fixed $\alpha$ is fine inside the actuator envelope; gating $\alpha$ on residual magnitude is left for future work. The language frontend is drop-in (Eq.~\ref{eq:grounding}); a stronger LLM or VLA can replace $\phi$ without retraining the controller. The TTA head is about 6.5K parameters and one forward pass per step, suitable for 50\,Hz real-time control, unlike gradient-based TTA which needs a backward pass each update~\cite{sun2020tta}. Everything in this paper is in simulation; real flights would also need sensor noise, communication delay, and actuator nonlinearity.

\section{Conclusion}
\label{sec:conclusion}

AeroBridge-TTA targets execution mismatch in language-conditioned UAV control with test-time adaptation. A single language-conditioned model does 5 tasks at $97.6\%$ average SR. Under 13 mismatch conditions with the same DR, it ties PPO-MLP in-distribution ($\Delta{=}0$) and wins every OOD condition ($+22.0$ pts, $7.7\times$ on Combined-OOD, $+33.3$ on Mass$+40\%$). The same-weights ablation isolates the mechanism: raising $\alpha$ from $0$ to $0.1$ lifts average OOD success from $16.8\%$ to $77.5\%$. The execution-mismatch view and this TTA formulation should carry over to other language-conditioned robots that face dynamics they did not train on.


\bibliographystyle{IEEEtran}
\bibliography{main}

\end{document}